\documentclass[letterpaper, 10 pt, conference]{ieeeconf}
\usepackage[T1]{fontenc}
\usepackage[latin9]{inputenc}
\usepackage{color}
\usepackage[dvipsnames]{xcolor}
\usepackage{float}
\usepackage{textcomp}
\usepackage{mathrsfs}
\usepackage{mathtools}
\usepackage{amsmath}

\usepackage{amsthm}
\usepackage{amssymb}
\usepackage{stmaryrd}
\usepackage{stackrel}
\usepackage{graphicx}
\usepackage{wasysym}
\usepackage{url}
\usepackage[font=small,labelfont=bf]{caption}
\usepackage{comment}
\usepackage{makecell}
\usepackage{multirow}
\usepackage{textgreek}
\usepackage{hyperref}
\makeatletter

\floatstyle{ruled}
\newfloat{algorithm}{tbp}{loa}
\providecommand{\algorithmname}{Algorithm}
\floatname{algorithm}{\protect\algorithmname}

\theoremstyle{plain}

\theoremstyle{definition}

\theoremstyle{definition}
\newtheorem{problem}{\protect\problemname}
\theoremstyle{plain}

\theoremstyle{definition}
\newtheorem{example}{\protect\examplename}
\theoremstyle{remark}

\theoremstyle{plain}

\providecommand{\lemmaname}{Lemma}
\usepackage{cite}
\usepackage{algpseudocode}
\usepackage{setspace}
\IEEEoverridecommandlockouts
\makeatother

\usepackage{babel}
\providecommand{\definitionname}{Definition}
\providecommand{\examplename}{Example}
\providecommand{\problemname}{Problem}
\providecommand{\theoremname}{Theorem}

\usepackage{color}

\begin{document}

\title{Exploiting Hybrid Policy in Reinforcement Learning for Interpretable
Temporal Logic Manipulation \thanks{H. Zhang, Hao Wang, and Z. Kan (Corresponding Author) are with the
Department of Automation at the University of Science and Technology
of China, Hefei, Anhui, China, 230026. X. Huang is with the School of Automation, Chongqing University, Chongqing, China. W. Chen is with School of Robotics, Hunan University, Changsha, China.} \thanks{This work was supported in part by National Key R\&D Program of China under Grant 2022YFB4701400/4701403 and National Natural Science Foundation of China under Grant U201360.}}

\author{Hao Zhang, Hao Wang, Xiucai Huang, Wenrui Chen, and Zhen Kan}
\maketitle
\begin{abstract}
Reinforcement Learning (RL) based methods have been increasingly explored for robot learning. However, RL based methods often suffer
from low sampling efficiency in the exploration phase, especially for 
long-horizon manipulation tasks, and generally neglect the semantic
information from the task level, resulted in a delayed convergence
or even tasks failure. To tackle these challenges, we propose a Temporal-Logic-guided Hybrid
policy framework (HyTL) which leverages three-level decision layers to improve the agent's performance. Specifically, the task specifications are encoded via linear temporal logic (LTL) to improve performance and offer interpretability. And a waypoints planning module is designed with the feedback from
the LTL-encoded task level as a high-level policy to improve the exploration efficiency. The middle-level policy selects which behavior primitives
to execute, and the low-level policy specifies the corresponding parameters to interact with
the environment. We evaluate HyTL on four challenging manipulation
tasks, which demonstrate its effectiveness and interpretability.
Our project is available at: \href{https://sites.google.com/view/hytl-0257/}{https://sites.google.com/view/hytl-0257/}.

\global\long\def\prog{\operatorname{prog}}
\global\long\def\argmax{\operatorname{argmax}}
\global\long\def\argmin{\operatorname{argmin}}
\end{abstract}

\section{Introduction}

A primary objective in robotic learning is to enable the robot to
automatically plan key points to accomplish a challenging
task like humans acting with instructions. To achieve such human-level intelligence, it is essential to understand the semantics of instructions and predict the
important states from the task feedback is crucial. Among numerous learning algorithms, reinforcement learning (RL) \cite{Sutton2018} exhibited strong potential in various applications \cite{jenelten2024dtc,cheng2023league,qiu2024instructing}.
While RL-based methods have empowered the agent to complete
tasks from simple to complex ones, a significant yet challenging
topic is how effectively the robots can plan key states as sub-goals
from the task level to ease the exploration burden and even improve the agent's performance for long-horizon tasks. In detail, there are three key challenges: 1) In contrast to traditional manipulation methods that learn from demonstration, how
can the robot plan the key states on its own to reduce the burden
on exploration? 2) When manipulating a long-horizon task, how to design an effective decision-making process to ease the exploration burden and incorporate the task semantics to facilitate the learning efficiency? 3) When considering long-horizon tasks,
how to interpret the robot motion planning in task
level?

Hierarchical reinforcement learning (HRL) that combines high-level
policy facilitating the accomplishment of long-horizon tasks and low-level control
policies has shown superior performance over conventional RL in a number of domains such as game scoring \cite{masson2016reinforcement} and \cite{fan2019hybrid}
as well as motion planning \cite{Wang2020Mobile} and \cite{shah2021value}.
When performing HRL in the field of robot manipulation, a hierarchical policy was developed in \cite{nasiriany2022augmenting} that chooses the action primitives and executes the corresponding parameters to accomplish challenging manipulation
tasks. A trajectory-centric smoothness
term is incorporated in \cite{fu2023metalearning} to enhance
the generalization in manipulation by using dynamic time warping to
align different trajectories to measure the distance. Empirical studies have found that the performance advantage of HRL
is mainly attributed to the use of sub-goal for augmenting exploration. However, many existing hierarchical
architectures are designed to be implemented directly at environment-level
manipulation, where the feedback of interaction determines the quality
of decision, lacking the connection with task semantics to guide
the robotic manipulations.

Owing to the rich expressivity and capability, linear temporal logic
(LTL) can describe a wide range of complex tasks combined by logically organized sub-tasks \cite{Belta2007,Baier2008,Cai2022}.
By transforming the LTL formula into an automaton, learning-based algorithms can be leveraged to solve manipulation tasks in robotic systems.
For example, a non-deterministic B\"uchi automaton is exploited in a high resolution grasp network (HRG-Net) \cite{Zhou2023Local} to facilitate reactive
human-robot collaboration in a locally observable transition system.
Truncated LTL is transformed to a finite-state predicate automaton
(FSPA) to facilitate the reward design and improve the performance
of manipulation in \cite{Li2019}. Similar to the automaton, the representation
module is also proposed by representing the LTL specification to guide the
agent for complex tasks \cite{Kuo2020}. To augment the representation ability, the work of
\cite{wang2023task} uses Transformer \cite{vaswani2017attention}
to express the semantics of LTL tasks for the robot's manipulation.
However, it cannot provide the agent with specific guidelines about how
to implement the LTL instructions from the task level to the concrete
environment. The work of \cite{xiong2023co} develops a hierarchical setting to
guide the robot to move in complex environments by waypoints. However,
its architecture makes it difficult to address long-horizon manipulations
and lacks the ability to provide interpretable guidance.

To bridge the gap, we consider planning the key points to ease the
exploration burden of the policy from the task level, and incorporate
the task semantics to provide meaningful interpretability for manipulations. The key contributions of this work are outlined below:

1) We develop a Temporal-Logic-guided Hybrid policy framework (HyTL),
which not only leverages a hybrid decision-making process to facilitate the
learning, but also incorporates task semantics to improve performance.

2) We design a novel waypoints planning module to ease the exploration
burden, which exploits the task feedback to guide the agent
interacting from the task level, empirically improving the agent's sampling
efficiency.

3) By leveraging gradients and disentangling features of the task representation
module, the interpretability of motion planning guided by LTL specifications
is further improved. 

4) We evaluate HyTL's performance on four challenging manipulation
tasks with five baselines, which exhibit higher learning efficiency,
better performance and interpretability compared to other methods
especially to \cite{wang2023task}.

\section{Preliminaries\label{sec:Pre}}

\subsection{Co-Safe LTL and LTL Progression}

Co-safe LTL (sc-LTL) is a subclass of LTL satisfied by
finite-horizon state trajectories \cite{Kupferman2001}. An sc-LTL formula is built on a set of atomic propositions
$\Pi$ that can be true or false, standard Boolean operators such
as $\wedge$ (conjunction), $\lor$ (disjunction), $\lnot$ (negation),
as well as temporal operators like  $\diamondsuit$ (eventually). The semantics of an sc-LTL formula are interpreted
over a word $\boldsymbol{\sigma}=\sigma_{0}\sigma_{1}...\sigma_{n}$,
which is a finite sequence with $\sigma_{i}\in2^{\Pi}$, where $i=0,\ldots,n$. 

LTL formulas can also be progressed along a sequence of truth assignments\cite{zhang2022temporal,Bacchus2000,tuli2022learning}.
Specifically, give an LTL formula $\varphi$ and a word $\boldsymbol{\sigma}=\sigma_{0}\sigma_{1}...$,
the LTL progression $\prog\left(\sigma_{i},\varphi\right)$ at step
$i$, $\forall i=0,1,...,$ is defined as $\prog\left(\sigma_{i},p\right)=\mathrm{True}\text{ if }p\in\sigma_{i}\text{, where }p\in\Pi$
and $\prog\left(\sigma_{i},p\right)=\mathrm{False}$ otherwise.

\subsection{Reinforcement Learning and Labeled PAMDP}

A parameterized action Markov decision process (PAMDP) \cite{masson2016reinforcement}
provides a primitive-augmented RL framework to address long-horizon tasks. The whole dynamics between the agent and environment over the sc-LTL tasks $\varphi$
can be modeled as a labeled PAMDP $\mathcal{M}_{e}=\left(S,T,\mathcal{H},p_{e},\Pi,L,R,\gamma,\mu\right)$,
where $S$ is the state space, $T$ indicates the horizon,   $\mathcal{H}=\left\{ \mathfrak{h}:\left(k,x_{k}\right)\mid x_{k}\in\mathcal{X}_{k}\text{ for all }k\in\mathcal{K}\right\} $
is the discrete-continuous
hybrid action space where $\mathcal{K}=\{1,...,K\}$ is the set of discrete behavior primitives
and $\mathcal{X}_{k}$ is the corresponding continuous parameter set
for each $k\in\mathcal{K}$,  $p_{e}(s'|s,\mathfrak{h})$ is the transition probability
from $s\in S$ to $s'\in S$ under action $\mathfrak{h}=\left(k,x_{k}\right)$, $\Pi$ is a set of atomic propositions, $L:S\rightarrow2^{\Pi}$ is the labeling function, $R:S\rightarrow\mathbb{R}$ is the
reward function, 
$\gamma\in\left(0,1\right]$ is the discount factor, and $\mu$ is the initial state distribution. A hybrid policy $\pi_{e}$ is exploited to interact with environment
under the task $\varphi$, which outputs a hybrid action pair $\mathfrak{h}$
and receives the corresponding reward by $r_{t}=R(s_{t})$.

\section{Problem Formulation\label{subsec:problem-formulation}}

In order to further elaborate the motivation of the proposed interpretable temporal-logic-guided hybrid
decision-making framework, we will use the following example throughout the work to illustrate the main idea of our method.
\begin{example}
\label{example1}
Consider a long-horizon manipulation skill of Peg Insertion from \cite{nasiriany2022augmenting}
as illustrated in Fig. \ref{fig:waypoint_plan_module}, in which the robot needs to grasp
the peg $O_{\mathsf{peg}}$ and inserts it into the hole $G_{\mathsf{hole}}$
guided by the planned waypoints. The set of propositions $\Pi$ is \{$\mathsf{peg\_grasped}$,
$\mathsf{hole\_reached}$, $\mathsf{peg\_inserted}$\}. Using above
propositions, an sc-LTL task is $\varphi_{\mathsf{peg}}=\lozenge(\mathsf{peg\_grasped}\wedge\lozenge(\mathsf{hole\_reached}\wedge\lozenge\mathsf{peg\_inserted}))$,
which requires the robot to sequentially grasp the peg, reach the hole and insert it into the hole.
\end{example}
In this work, we are interested in designing a temporal-logic guided hybrid policy architecture, which not only plans key waypoints based
on the task semantics, but also guides the agent through the waypoints
to choose appropriate behavioral primitives and the corresponding
parameters to facilitate robot learning. By predicting the
key states via the planning module, we hope to take advantage of its
foresight to generate hypothetical goals to guide the agent, and ease
the exploration burden of the robot\textquoteright s motion planning.
Compared to hierarchical architectures \cite{nasiriany2022augmenting} that are committed to manipulation,
the waypoints generated by the planning module are gradually updated
according to LTL instructions to guide the agent towards  sub-goals
as quickly as possible, resulting in a three-way improvement, in which the
LTL instructions and environmental rewards boost the planning module
for better waypoints, the waypoints guide the robot's manipulation, and the output actions improve Transformer for better tasks semantics.

By exploiting the LTL progression, an augmented PAMDP with an LTL
instruction $\varphi$, namely the task-driven labeled PAMDP (TL--PAMDP),
is developed as $\mathcal{M}_{\mathcal{H}}^{\varphi}\triangleq\left(\tilde{S},\tilde{T},\mathcal{H},\tilde{p},\Pi,L,\tilde{R}_{\Psi},\gamma,\mu\right)$
like \cite{wang2023task} (Detailed definitions can be found on our \href{https://sites.google.com/view/hytl-0257/} {website}), where the reward function is
\begin{equation}
\tilde{R}_{\varphi}(s,\varphi)=\begin{cases}
r_{env}+r_{\varphi}, & \text{if \ensuremath{\mathrm{prog}(L(s),\varphi)=\mathrm{True}}}\\
r_{env}-r_{\varphi}, & \text{if \ensuremath{\mathrm{prog}(L(s),\varphi)=\mathrm{False}}}.\\
r_{env}, & otherwise
\end{cases}\label{eq:Markovian Reward}
\end{equation}

The problem of this work can be stated as follows.

\begin{problem}
\label{Prob1}Given a TL--PAMDP $\mathcal{M}_{\mathcal{H}}^{\varphi}\triangleq\left(\tilde{S},\tilde{T},\mathcal{H},\tilde{p},\Pi,L,\tilde{R}_{\Psi},\gamma,\mu\right)$
corresponding to task $\varphi$,
this work is aimed at finding an optimal policy $\pi_{\mathcal{\varphi}}^{*}$
over the LTL instruction $\varphi$, so that the return $\mathbb{E}\left[\stackrel[k=0]{\infty}{\sum}\gamma^{k}r_{t+k}\mid S_{t}=s\right]$
under the policy $\pi_{\mathcal{\varphi}}^{*}$ can be maximized.
\end{problem}

\section{Algorithm Design\label{sec:Algorithm Design}}

To address Problem \ref{Prob1}, we present a novel approach, namely
\textbf{T}emporal\textbf{-L}ogic-guided \textbf{Hy}brid policy framework
(HyTL), that offers a hybrid policy to plan waypoints, choose primitives,
and determine parameters based on LTL representation encoded by Transformer.
Section \ref{subsec:plan waypoints} presents how the plan module
in HyTL generates hypothetical waypoints to improve sampling efficiency. Section \ref{subsec:Hyprid Policies} explains
in detail how hybrid decision-making architecture facilitates
the agent's training in long-horizon manipulations. Section \ref{subsec:AttCAT}
shows how AttCAT interprets the LTL instruction
for the robot's manipulation. The overall method of HyTL is illustrated
in Fig. \ref{fig:Framework}, which first extracts the LTL representation
$\varphi_{\theta}$ by the task representation module, then samples
a series of waypoints $\boldsymbol{w}$ based on the initial state
$s_{0}$ by the waypoints planning module, and outputs the action
pair $\mathfrak{h}=\left(k,x_{k}\right)$ by selecting the appropriate
action primitive $k$ following the primitives choosing module and
determines the corresponding parameters $x_{k}$ following the\textbf{
}parameterization determining module.

\subsection{Plan Waypoints from Task Feedback\label{subsec:plan waypoints}}

One of the main challenges in solving Problem \ref{Prob1} is sampling
efficiency in the exploration phase when utilizing RL for long-horizon
manipulations. To address this problem, \cite{wang2023task} considers
augmenting the hierarchical policy with task semantics to enable agent
making decisions corresponding to the sub-goal and accelerate learning.
However, it is difficult for this approach to map abstract task semantics
into a concrete manipulation. Therefore, we design a planning module
that incorporates task feedback to generate waypoints to guide the
agent by combining task rewards.

\begin{figure}
\centering{}\includegraphics[scale=0.25]{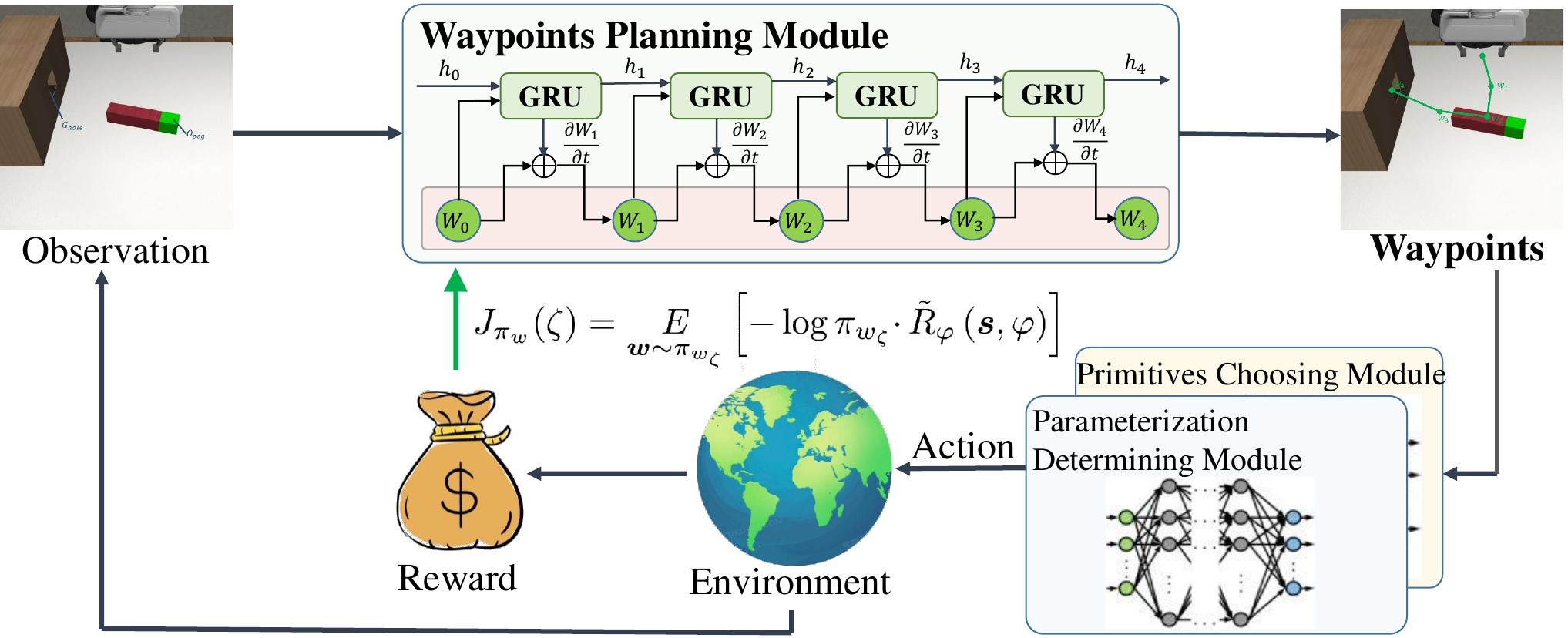}\caption{\label{fig:waypoint_plan_module} The waypoints planning
module updated via interactions.}
\end{figure}

\textbf{Waypoints Planning Module}. As shown in Fig. \ref{fig:waypoint_plan_module},
we exploit the Gated Recurrent Unit (GRU) as a predictor in the form of a residual connection to incrementally
construct waypoints to guide agents. Specifically, given an initial
state $s_{0}$ as the initial waypoints $w_{0}$ and hidden state
$h_{0}$, the sequential waypoints $\boldsymbol{w=}w_{0}w_{1}...$
can be generated by the deviation between subgoals as 
$
w_{i+1}=w_{i}+\frac{\partial w_{i}}{\partial t}
$
where $\mathrm{\frac{\partial w_{i}}{\partial t}=GRU}\left(w_{i},h_{i}\right)$ and $w_{i}\in\mathbb{R}^{3}$.
Since it's hard for GRU to update directly
through rewards, we model the plan module as a stochastic policy with
Gaussian distribution inspired by \cite{xiong2023co}, i.e., $\pi_{w_{\zeta}}=\mathcal{N}(\overline{w},\sigma)$ with weights $\zeta$
where $\overline{w}$ is the mean of waypoint $w_{i}$ and $\sigma$
is predicted by a linear transform from the hidden state $h_{i}$.
Thus a sequential waypoints $\boldsymbol{w=}w_{0}w_{1}...$ can be
sampled from $\pi_{w_{\zeta}}(s_{0})$ and if the agent reaches $w_{i}$,
$w_{i+1}$ will be exploited to guide. To bridge the environment and the feedback of the task (i.e., satisfy the
LTL task $\varphi$), we design the following loss function 
\[
J_{\pi_{w}}(\zeta)=\underset{\boldsymbol{w}\sim\pi_{w_{\zeta}}}{E}\left[-\log\pi_{w_{\zeta}}\cdotp\tilde{R}_{\varphi}\left(\boldsymbol{s},\varphi\right)\right]
\]
to update
the planning module, where $\tilde{R}_{\varphi}\left(\boldsymbol{s},\varphi\right)$ is
the cumulative reward from the state sequence $\boldsymbol{s=}s_{0}s_{1}...$ corresponding to $\boldsymbol{w}$.
Thus when the planning layer is
updated, it not only uses the environment-level reward $r_{env}$,
but also exploits the task-level reward $r_{\varphi}$ to continuously
optimize the waypoints. 

As shown in Fig. \ref{fig:Framework}(c), by using the waypoint planning module as a high-level policy layer,
not only can the task-level semantics be taken over to guide the agent
at the environment, but also the residual structure can be exploited
to provide the agent with more flexible guidance by setting up more
waypoints in hard-to-explore areas.

\subsection{Hybrid Policy based on LTL Instructions\label{subsec:Hyprid Policies}}

\begin{figure}
\centering{}\includegraphics[scale=0.245]{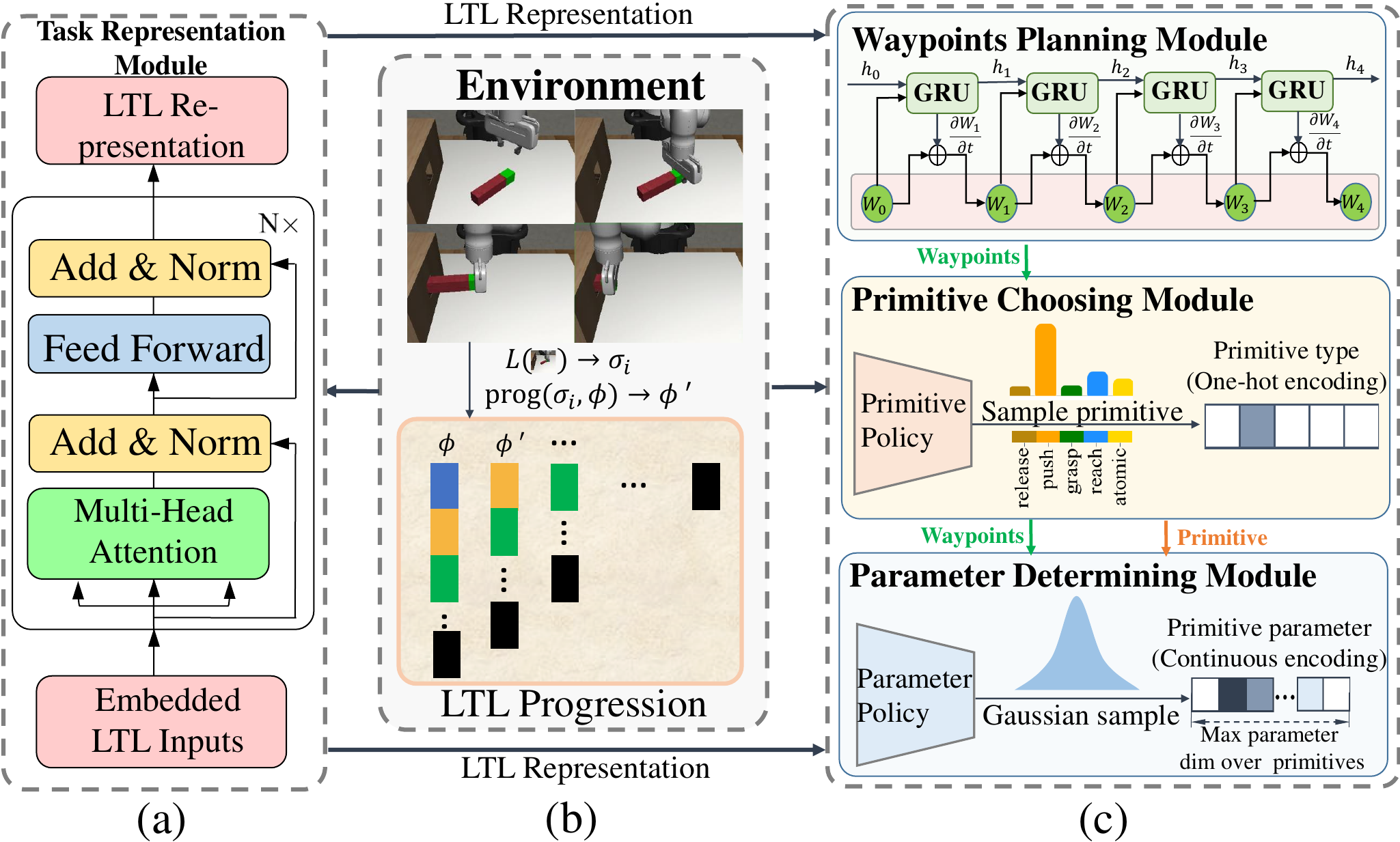}\caption{\label{fig:Framework}The framework of HyTL. (a) The outline of Task Representation Module. (b) The
LTL progression for progressing LTL formulas. (c) The hybrid decision-making
process.}
\end{figure}

Another challenge in solving Problem 1 is to design an effective
decision-making architecture to improve the agent's performance in long-horizon
manipulations. To address this challenge, the work of \cite{nasiriany2022augmenting} pre-built five behavioral primitives (\textbf{reach}, \textbf{grasp},
\textbf{push}, \textbf{release}, and \textbf{atomic}), and designed
a hierarchical architecture to reduce the agent's exploration
burden by first selecting appropriate primitives through a high-level
policy and then determining the parameters of primitives via low-level
policies. However, \cite{nasiriany2022augmenting} focuses on the
design of the decision structure, neglecting how task-level information
can help the agent accomplish the task. To address this problem,
based on LTL representation, this work proposes a hybrid decision-making
framework that not only incorporates task semantics to improve the
learning efficiency, but also plans waypoints from the task level
to guide the agent to accomplish long-horizon tasks. 
Except for Waypoints Planning Module mentioned in Section \ref{subsec:plan waypoints}, the HyTL framework shown in Fig .\ref{fig:Framework} comprises following key modules.

\textbf{Task Representation Module.} To incorporate the task semantics, we use Transfomer \cite{vaswani2017attention}
to encode LTL representation. Given an input $X_{\varphi}=(x_{0},x_{1},...)$
generated by the LTL task $\varphi$ where $x_{t,t=0,1,...},$ represents
the operator or proposition, $X_{\varphi}$ is preprocessed using
the word embedding $\mathrm{E}$ as $X_{\mathrm{E}}=\left[x_{0}\mathrm{E};x_{1}\mathrm{E};...;x_{N}\mathrm{E}\right]\in\mathbb{R}^{{\normalcolor B\times M\times}D}$
where $B$ is the batch size, $M=N+1$ is the length of input $X_{\varphi}$
and $D$ is the model dimension of Transformer. $X_{\mathrm{E}}$
is then combined with the frequency-based positional embedding $E_{pos}\in\mathbb{R}^{B\times M\times D}$
to utilize the sequence order. Then the LTL representations encoded by Transformer can be computed by following steps:
\begin{equation}
\begin{array}{cc}
X_{0}=X_{\mathrm{E}}+E_{pos},\\
X_{l}^{'}=\text{MSA}(\text{LN}(X_{l-1}))+X_{l-1}, & l=1,...,L\\
X_{l}=\text{MLP}(\text{LN}(X_{l}^{'}))+X_{l}^{'}, & l=1,...,L\\
Y=\text{LN}(X_{l})
\end{array}\label{eq:Encoder4Trans}
\end{equation}
where MSA denotes the multi-head self-attention, LN means the layer norm, MLP represents the position-wise fully connected feed-forward sub-layers, and $Y$ is the output of the final layer from the Transformer
encoder. The outline of Transformer Encoder for HyTL is shown
in Fig. \ref{fig:Framework}(a). Thus incorporating the LTL representation
into HyTL, which not only helps to improve the agent's performance,
but also lays the foundation for the interpretability of motion planning
shown in Section \ref{subsec:AttCAT}.

\textbf{Primitives Choosing Module}. Based on the predicted waypoint
$w_{i}$ as guidance, the primitive policy outputs a behavioral primitive appropriate for the current state and the progressed LTL instruction.
Specifically, the behavior primitive $k$ can be chosen by primitive
policy $\pi_{k_{\psi}}(k|s,\varphi_{\theta},w)$ with weights $\psi$ conditioning on the waypoint $w$. As a middle-level policy layer, the primitives choosing
module not only considers the information
from the task level and environment to offer the suitable primitive,
but also guides the low-level parameter module on what to do for long-horizon
manipulations. 

\textbf{Parameterization Determining Module}. Conditioning on the
current state, task representations, predicted waypoints, and selected
primitives, the parameterized policy outputs an appropriate set of
parameters to ensure effective interactions between the behavioral primitives and the environment. Specifically, the primitive parameter
$x$ can be determined by parameter policy $\pi_{p_{\xi}}(x|s,\varphi_{\theta},w,k)$
with weights $\xi$ conditioning on the waypoint $w$ and primitive
$k$. As the last layer interacts with the environment, the parameterization
determining module, based on the above conditions, instructs the agent
on how to interact with the environment to accomplish the tasks.

In this work, we opt for SAC \cite{haarnoja2018soft} as the RL backbone
to update in the framework. Let $Q_{\varphi_{\theta}}(s,\mathfrak{h},w)$
and $Q_{\varphi_{\theta}^{'}}(s,\mathfrak{h},w)$ be the Q-value function
of task $\varphi$ and $\varphi^{'}$, and let $\pi_{w_{\zeta}}(s)$,
$\pi_{k_{\psi}}(k|s,\varphi_{\theta},w)$ and $\pi_{p_{\xi}}(x|s,\varphi_{\theta},w,k)$
be the hybrid policy networks. The loss for critic, plan policy, primitive
policy, and parameter policy in adapted SAC are then designed respectively
as
\begin{equation}
\begin{array}{c}
J_{Q}=(Q_{\varphi_{\theta}}-(\tilde{R}_{\varphi}+\gamma(Q_{\varphi_{\theta}^{'}}-\alpha_{k}\log\pi_{k_{\psi}}-\alpha_{p}\log\pi_{p_{\xi}})))^{2},\\
J_{\pi_{w}}(\zeta)=\underset{\boldsymbol{w}\sim\pi_{w_{\zeta}}}{E}\left[-\log\pi_{w_{\zeta}}\cdotp\tilde{R}_{\varphi}\left(\boldsymbol{s},\varphi\right)\right],\\
J_{\pi_{k}}(\psi)=\underset{\boldsymbol{w}\sim\pi_{w_{\zeta}}}{E}\underset{k\sim\pi_{k_{\psi}}}{E}\left[\alpha_{k}\log\pi_{k_{\psi}}-\underset{x\sim\pi_{p_{\xi}}}{E}\left[Q_{\varphi_{\theta}}\right]\right],\\
J_{\pi_{p}}(\xi)=\underset{\boldsymbol{w}\sim\pi_{w_{\zeta}}}{E}\underset{k\sim\pi_{k_{\psi}}}{E}\underset{x\sim\pi_{p_{\xi}}}{E}\left[\alpha_{p}\log\pi_{p_{\xi}}-Q_{\varphi_{\theta}}\right].
\end{array}\label{eq:update equation}
\end{equation}
 Note that $\varphi_{\theta}$ and $\varphi_{\theta}^{'}$ encoded
by Transformer are indirectly updated by back-propagation of the above
equations. 

The pseudo-code is outlined in Alg. \ref{Alg_HyTL}. In
the exploration phase, Transformer first extracts the LTL representation
$\varphi_{\theta}$ and concatenates the valve with the state $s_{0}$
(line 4). Then before the agent interacts with the environment, the
waypoint planning module samples a series of waypoints $\boldsymbol{w}$
based on the initial state $s_{0}$ and incorporates $w_{i}$ as part
of the observation to guide the agent's exploration (line 6). Based
on the above observation, the agent selects the appropriate action
primitive following $\pi_{k_{\psi}}$ and determines the corresponding
parameters by $\pi_{p_{\xi}}$ (line 11). During the interaction,
the operator $\mathrm{prog}$ tracks the original instructions
$\varphi$ and checks whether the LTL task
$\varphi$ is complete (lines 7-10). In the training phase, HyTL updates
all neural network weights following (\ref{eq:update equation}) (lines
14-16).

\subsection{Interpret Manipulation via AttCAT\label{subsec:AttCAT}}

Another advantage of representing LTL instructions with Transformer
is that it provides interpretability for robot motion planning. When
LTL specifications are encoded via Transformer \cite{zhang2023exploiting},
their interpretability can be represented by the value of heads'
weights on different propositions. However, \cite{zhang2023exploiting}
only sums the weights of all heads, ignoring the effect of the gradients
flowing in the Transformer architecture. Inspired by \cite{qiang2022attcat},
this work further explores the impact of LTL representation
on robot motion planning by interpreting Transformer via Attentive
Class Activation Tokens (AttCAT), which not only utilizes the inputs' gradients combined with attention weights to generate impact scores, but also
disentangles features flowing between intermediate layers of Transformer.

\begin{algorithm}
\caption{\label{Alg_HyTL}Temporal\textbf{-}Logic-guided Hybrid policy (HyTL)}

\scriptsize

\singlespacing

\begin{algorithmic}[1]

\Procedure {Input:} {The PAMDP $\mathcal{M}_{e}$ with the LTL specification $\varphi$}

{Output: } { An approximately optimal policy $\pi_{\varphi}^{*}(\mathfrak{h}\mid s,\varphi)$
for the TL-PAMDP $\mathcal{M}_{\mathcal{H}}^{\varphi}$}

{Initialization: } { All neural network weights}

\For {iteration 1,2,...,N} \{Exploration Phase\}

\For {episode 1,2,...,M}

\State Initialize timer $t\leftarrow0$ and episode $s_{0}$, and
augment the state $s_{0}$ with $\varphi_{\theta}$ encoded by Transformer

\While {episode not terminated}

\State Sample waypoints $\boldsymbol{w=}w_{0}w_{1}...$ from $\pi_{w_{\zeta}}(s_{0})$
and guide the state $s_{t}$ by the waypoint $w_{i}$

\State $\varphi^{'}\leftarrow\mathrm{prog}(L(s),\varphi)$

\If {$\varphi^{'}\in\{\mathsf{True,\mathsf{False}}\}$ or $s\in T$
}

\State Break

\EndIf

\State Gather data from $\varphi$ following $\pi_{k_{\psi}}$ and
$\pi_{p_{\xi}}$, and guide the state $s_{t}$ by the waypoint $w_{i+1}$
if the agent reached $w_{i}$

\EndWhile

\EndFor

\For {training step 1,2,...,K} \{Training Phase\}

\State Update all neural network weights by (\ref{eq:update equation})

\EndFor

\EndFor

\EndProcedure

\end{algorithmic}
\end{algorithm}

Based on (\ref{eq:Encoder4Trans}),
we can write columns of $X_{l}$ separately as $\mathfrak{x}_{i}^{l}$,
$i=0,1,...,M$. To analyze the effect of $i$-th token on the output
$y^{c}$ across different proposition class $c$ in the embedded LTL
input $X_{0}$, the relationship between $y^{c}$ and $\mathfrak{x}_{i}^{L}$
can be modeled by a linear relationship 
$
y^{c}\sideset{=}{_{i}^{N+1}}\sum w_{i}^{c}\cdot\mathfrak{x}_{i}^{L},
$
where $w_{i}^{c}=\frac{\partial y^{c}}{\partial\mathfrak{x}_{i}^{L}}$
is the linear coefficient vector of $\mathfrak{x}_{i}^{L}$ capturing
the impact of $i$-th token on the target class $c.$

By considering the interaction among tokens, the impact score of the
$i$-th token towards class $c$, denoted by AttCAT, can be formulated
through the weighted combination: 
\[
\mathrm{AttCAT_{i}}\sideset{=}{_{l=1}^{L}}\sum E_{H}(\alpha_{\text{i}}^{l}\cdot(w_{i}^{c}\odot\mathfrak{x}_{i}^{l})),
\]
where $\odot$ denotes the Hadamard product, $\alpha_{\text{i}}^{l}$
represents the attention weights of the $i$-th token at $l$-th
layer, and $E_{H}(\cdot)$ stands for the mean over multiple heads. 

When provided with a pre-trained Transformer $TF(\cdot)$, an input LTL instruction
$\varphi$, and the interpretability method $\mathrm{AttCAT}(\cdot)$,
the magnitude of impact can be calculated by $\left|\mathrm{AttCAT}(TF(\varphi))\right|$,
reflecting each token's contribution.
By the visualization input tokens scores, the most impactful token on the output can be identified. 

\section{CASE STUDIES}

In this section, the performance of the HyTL framework is evaluated
in comparison to state-of-the-art algorithms through scenarios. We
specifically represent the following aspects. \textbf{1) Performance:}
how effectively does HyTL surpass the state-of-the-art
algorithm in long-horizon manipulations? \textbf{2) Architecture:
}How good is the waypoints planning module for algorithmic
enhancement? \textbf{3) Interpretable: }How well can the agent comprehend
the LTL instruction?

\subsection{Baselines and Tasks Setting}

\textbf{Baselines. }To demonstrate the efficacy of the HyTL framework, we empirically compare its performance against five baselines. 1) Our RL backbone \textbf{SAC} from \cite{haarnoja2018soft} is the first baseline, which only executes
atomic primitive. 2) The second baseline is \textbf{Maple
} from \cite{nasiriany2022augmenting}
which augments the traditional RL methods with action primitives and corresponding parameters to improve the exploration efficiency. 3) The third baseline
is \textbf{$\boldsymbol{\mathrm{Maple_{way}}}$}, which is based on Maple and exploits the planning
module to ease the exploration burden. 4) The fourth baseline is \textbf{$\boldsymbol{\mathrm{Maple_{LTL2Action}}}$},which
further exploits the encoder from \cite{vaezipoor2021ltl2action}
to represent semantics of LTL for improving sampling efficiency. 5)
The fifth baseline is \textbf{$\boldsymbol{\mathrm{TARPs}\mathrm{_{TF-LTL}}}$},
which is a manipulation skill learning algorithm from \cite{wang2023task}
augmenting RL with temporal logic and hybrid action space.

\textbf{Tasks Setting. }To evaluate the algorithm performances, four challenging manipulations in \cite{nasiriany2022augmenting}
are employed, including \textbf{Stack}, \textbf{Nut Assembly},
\textbf{Cleanup} and \textbf{Peg Insertion}. The corresponding
task descriptions and LTL instructions are stated in
Table. II of \cite{wang2023task} and Example. \ref{example1}.

\subsection{Experimental Results and Analysis}

\begin{figure}
\centering{}\includegraphics[scale=0.29]{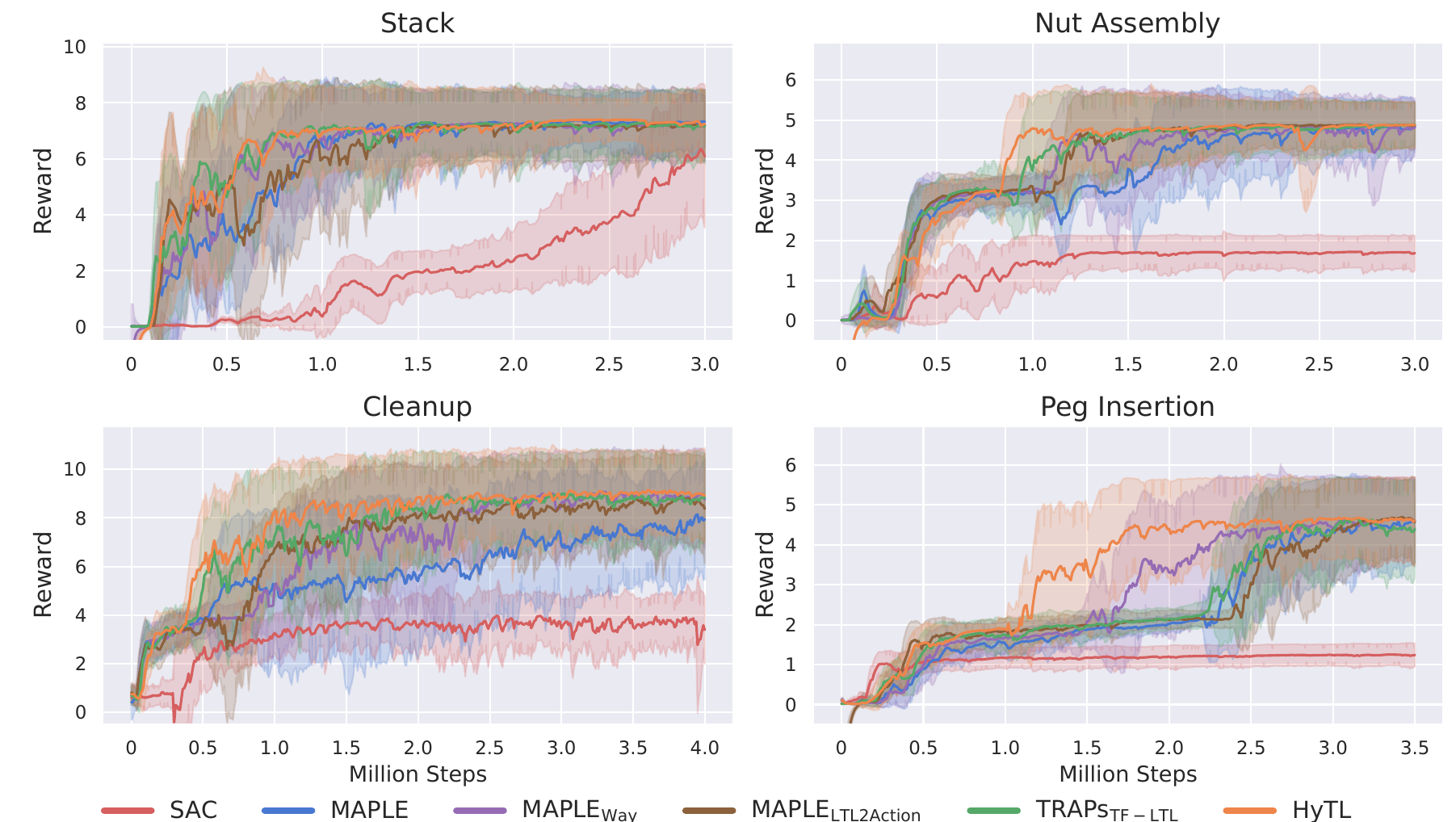}\caption{\label{fig:performance}Plots of normalized reward curves for four manipulations.}
\end{figure}

\textbf{(1) Main Results.} Fig. \ref{fig:performance} illustrates the
results performed by different approaches over 6 seeds. As shown in
Fig. \ref{fig:performance}, it is observed that 1) algorithms guided
by waypoints ($\mathrm{Maple_{way}}$ and HyTL) are more efficiently
sampled and converge faster than those without waypoints (Maple,
$\mathrm{Maple_{LTL2Action}}$, and $\mathrm{TARPs}\mathrm{_{TF-LTL}}$);
2) HyTL exhibits better performance relative to $\mathrm{TARPs}\mathrm{_{TF-LTL}}$
and $\mathrm{Maple_{LTL2Action}}$ which only incorporate the task semantics; 3) On the most challenging task Peg Insertion, HyTL demonstrates the shortest converge episode, with over
30\% reduction compared to the best previous work $\mathrm{TARPs}\mathrm{_{TF-LTL}}$ \cite{wang2023task}. 

From the above observation, we conjecture that the unstable representation
of the task module at the beginning of training affects the performance
of the agents, which rely heavily on the task representation to improve
the sampling efficiency. In contrast, the performance of HyTL, which
has a hybrid decision architecture design, is not only affected by
the task module's representation, but also relies on the waypoints
generated by the planning module for guidance. Once either the planning
module or the task module has been effectively updated, the gradient decreases rapidly in the direction that contributes
to task completion. It is due to this complementary design in the
hybrid decision architecture that HyTL can demonstrate higher learning
efficiency.

\begin{figure}
\centering{}\includegraphics[scale=0.18]{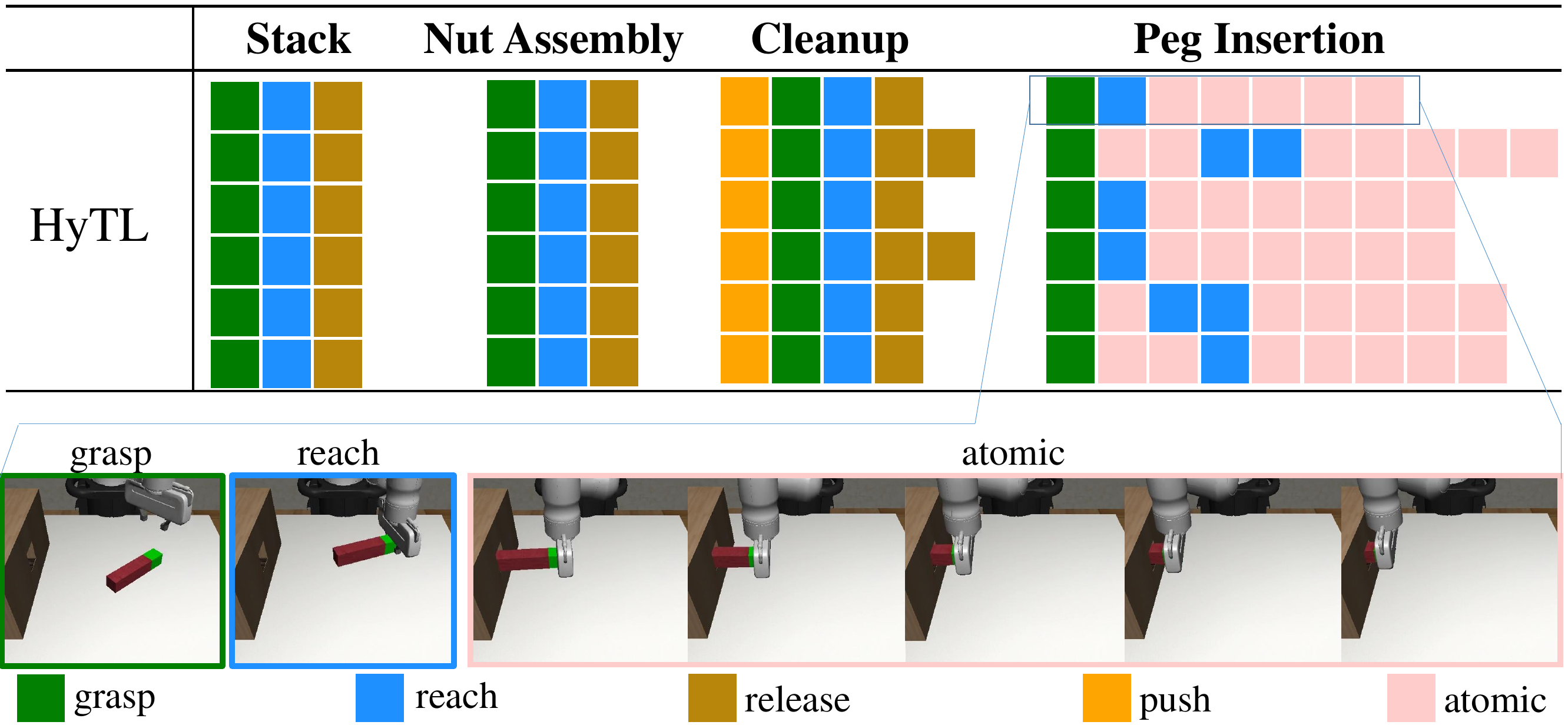}\caption{\label{fig:HyTL_sketches} The visualization of action sketches utilizing HyTL.}
\end{figure}

\textbf{2) Primitive Compositionality Quantification. }Since primitives
are utilized in HyTL like Maple, we use the compositionality
score from \cite{nasiriany2022augmenting} to evaluate the degree
of action primitives over different methods. The action sketches of HyTL with 6 seeds are visualized in Fig. \ref{fig:HyTL_sketches},
which shows the different action primitives that HyTL selects and
combines in accomplishing above four manipulation tasks. The compositionality
scores are shown in Table \ref{tab:Compositionality Score}, in which higher
scores reflect better compositionality and more stable performance.
As shown in Table \ref{tab:Compositionality Score}, $\mathrm{Maple_{way}}$
and HyTL can select and combine more appropriate behavior primitives
by guiding within waypoints, resulting in higher compositionality
scores than other methods.

\begin{figure}
\centering{}\includegraphics[scale=0.20]{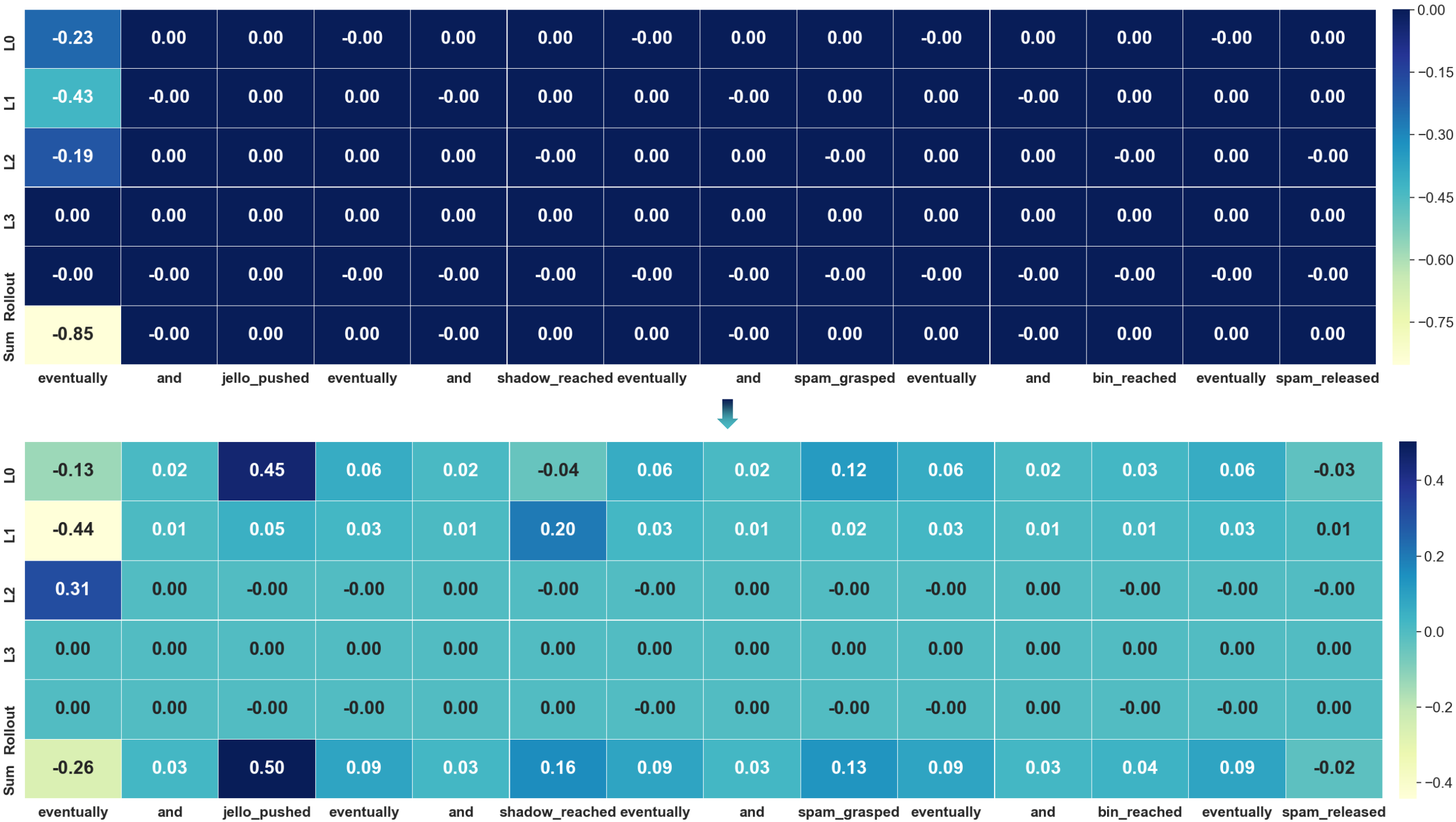}\caption{\label{fig:interpret via AttCAT} We illustrate the heatmap of the task $\varphi_{\mathsf{cleanup}}$ by normalizing impact scores from different Transformer layers.}
\end{figure}

\textbf{(3) Interpretability via AttCAT. }To further illustrate the agent's understanding of the LTL task,
Fig. \ref{fig:interpret via AttCAT} illustrates the heatmap of the task $\varphi_{\mathsf{cleanup}}$. As shown in the top table of Fig. \ref{fig:interpret via AttCAT}, the
cumulative scores for all tokens except for the $\mathsf{eventually}$(-0.85)
are almost zero, indicating that the agent lacks a clear understanding of the LTL instruction at the beginning of the training. Upon the convergence of Transforme, higher impact scores from all layers focus
on the token $\mathsf{jello\_pushed}$(+0.50) as shown in the bottom
row of Fig. \ref{fig:interpret via AttCAT}, which suggests that the agent is more likely to move directly to the position corresponding to the $\mathsf{jello\_pushed}$ proposition. 



\begin{table}
\caption{\label{tab:Compositionality Score} We present compositionality scores to reflect the compositionality and stability of the algorithms in four scenarios.}

\centering{}\resizebox{0.47\textwidth}{!}{
\begin{tabular}{c|ccccc}
\hline 
Compositionality Score & Maple & $\mathrm{Maple_{way}}$ & $\mathrm{Maple_{LTL2Action}}$ & $\mathrm{TARPs}\mathrm{_{TF-LTL}}$ & \multicolumn{1}{c}{HyTL}\tabularnewline
\hline 
Stack & 0.71 & 1.00 & 0.79 & 0.92 & \textbf{1.00}\tabularnewline
Nut Assembly & 0.75 & 1.00 & 0.85 & 1.00 & \textbf{1.00}\tabularnewline
\multirow{1}{*}{Cleanup} & 0.73 & 0.84 & 0.75 & 0.83 & \multicolumn{1}{c}{\textbf{0.89}}\tabularnewline
Peg Insertion & 0.24 & \textbf{0.91} & 0.83 & 0.81 & 0.86\tabularnewline
\hline 
\end{tabular}}
\end{table}

\section{CONCLUSIONS}

In this work, we present an interpretable temporal-logic-guided hybrid
decision-making framework to improve the agent's performance on four challenging
manipulation tasks. In particular, a novel waypoints planning module
is designed to ease the exploration burden, which exploits the task
feedback to guide the agent interacting from the task level. And the
hybrid decision-making process with three-level decision layers is
proposed to facilitate learning of agent's manipulations. In addition,
the interpretability of motion planning guided by LTL specifications
is further improved by leveraging gradients and disentangling features
of the task representation module. Experimental results and analysis
demonstrate that HyTL improves the agent\textquoteright s sampling
efficiency and offers reasonable interpretability. Future work will
consider extending the method of HyTL to more challenging tasks, such
as dexterous manipulations.

\bibliographystyle{IEEEtran}
\bibliography{main}

\end{document}